\documentclass[10pt,twocolumn,letterpaper]{article}
\usepackage{graphicx}
\usepackage{amsmath}
\usepackage{amssymb}
\usepackage{booktabs}
\usepackage{enumitem}
\usepackage{indentfirst}
\usepackage{multirow}
\usepackage{multicol}
\usepackage[accsupp]{axessibility}
\usepackage{float}
\usepackage{bbding}
\usepackage{amsmath}

\usepackage[pagenumbers]{cvpr} 

\usepackage[dvipsnames]{xcolor}

\definecolor{cvprblue}{rgb}{0.21,0.49,0.74}
\usepackage[pagebackref,breaklinks,colorlinks,citecolor=cvprblue]{hyperref}
\usepackage{pifont}
\definecolor{limegreen}{HTML}{32CD32}
\newcommand{\cmark}{\textcolor{limegreen}{\ding{51}}}%
\newcommand{\xmark}{\textcolor{red}{\ding{55}}}%

\def\paperID{3063} 
\def\confName{CVPR}
\def\confYear{2025}
\def\Methodname{UMGen}

\title{Generating Multimodal Driving Scenes via Next-Scene Prediction}
\author{
Yanhao Wu$^{1, 2}$ {\textsuperscript{*}} ~
Haoyang Zhang$^{2}$ ~
Tianwei Lin$^{2}$ ~
Lichao Huang$^{2}$ ~
Shujie Luo$^{2}$ \\
Rui Wu$^{2}$ ~
Congpei Qiu$^{1}$ ~
Wei Ke$^{1}$\href{mailto:wei.ke@mail.xjtu.edu.cn}{\textsuperscript{†}} ~ 
Tong Zhang$^{3, 4}$ \\
$^1$ School of Software Engineering, XJTU ~
$^2$ Horizon Robotics \\
$^3$ School of Computer and Communication Sciences, EPFL  \\
$^4$ University of Chinese Academy of Sciences\\
}

\usepackage{xcolor}

\usepackage[utf8]{inputenc}
\DeclareUnicodeCharacter{FF0C}{,}

\begin{document}

\twocolumn[{%
\renewcommand \twocolumn[2][]{#1}%
\maketitle
\vspace{-2em}
\includegraphics[width=\textwidth]{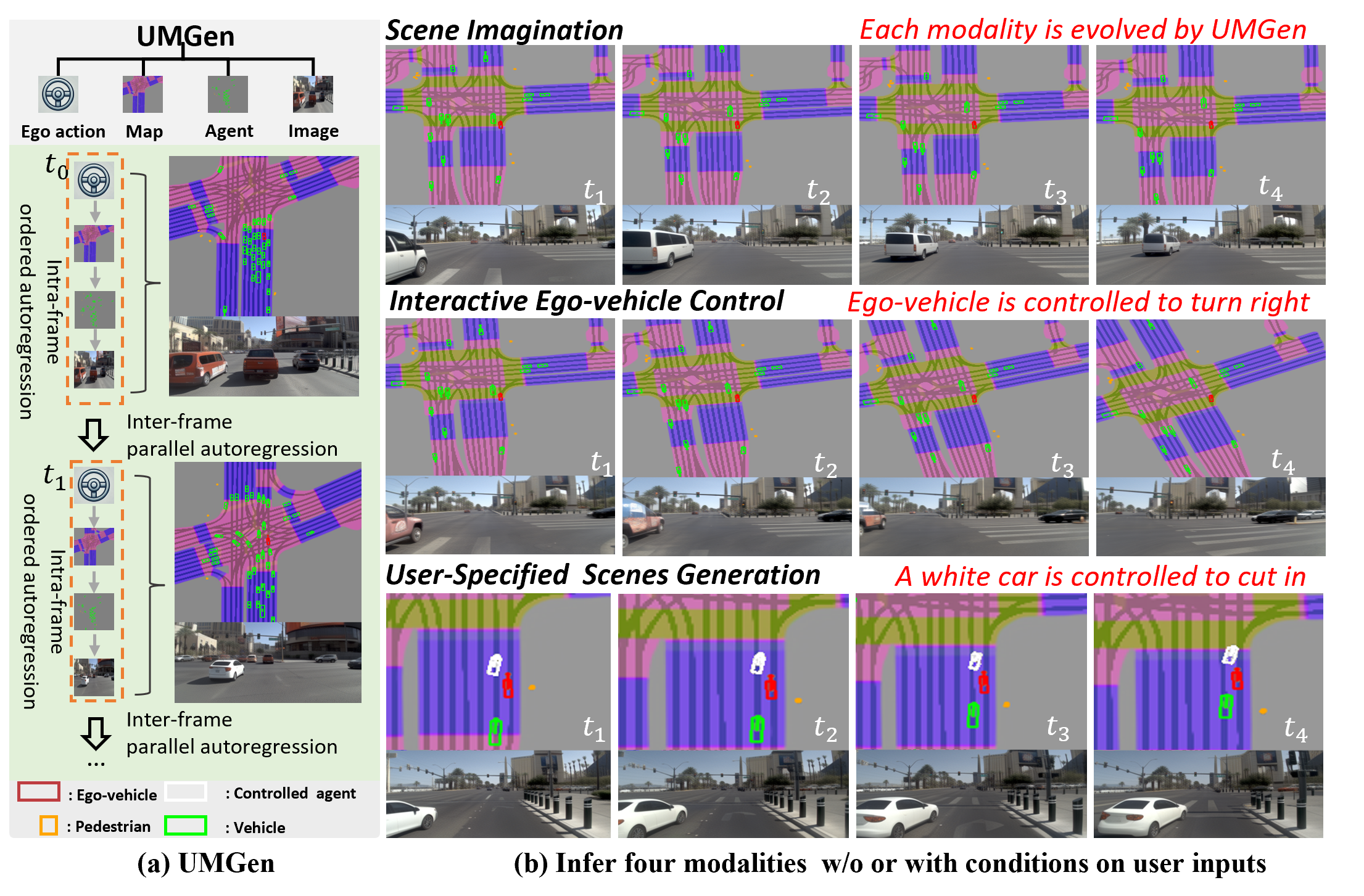}
\vspace{-2.5em}
\captionof{figure}{ An overview of our proposed driving scene generation paradigm $-$ \Methodname. Starting from a random initialization (a) \Methodname~ generates ego-centric, multimodal scenes frame-by-frame. Each scene encompasses four modalities: ego-vehicle action, map, traffic agent, and image; (b) \Methodname~offers multiple functions. It can autonomously generate multimodal scene sequences based solely on its own historical context,  but also predict the other modalities based on input ego-vehicle actions provided by users. Furthermore, \Methodname~can incorporate user-specified agent actions to create customized scene sequences. In three scene sequences, arranged from top to bottom, we demonstrate the ego vehicle autonomously driving straight through an intersection, executing a user-defined right turn, and encountering scenes where a user-specified white car cuts in front of it.
For better visualization, a portion of the map corresponding to the user-specified scenario is zoomed in.}
\vspace{0.75em}
\label{fig:Teaser_Big}
}]

\begin{abstract}
\vspace{-2em}
Generative models in Autonomous Driving (AD) enable diverse scenario creation, yet existing methods fall short by only capturing a limited range of modalities, restricting the capability of generating controllable scenes for comprehensive evaluation of AD systems. In this paper, we introduce a multimodal generation framework that incorporates four major data modalities, including a novel addition of the map modality. With tokenized modalities, our scene sequence generation framework autoregressively predicts each scene while managing computational demands through a two-stage approach. The Temporal AutoRegressive (TAR) component captures inter-frame dynamics for each modality, while the Ordered AutoRegressive (OAR) component aligns modalities within each scene by sequentially predicting tokens in a fixed order. To maintain coherence between map and ego-action modalities, we introduce the Action-aware Map Alignment (AMA) module, which applies a transformation based on the ego-action to maintain coherence between these two modalities.  Our framework effectively generates complex, realistic driving scenes over extended sequences, ensuring multimodal consistency and offering fine-grained control over scene elements.
The project page can be found at: \href{https://yanhaowu.github.io/UMGen/}{https://yanhaowu.github.io/UMGen}.
\renewcommand{\thefootnote}{} 
\footnote{%
\parbox{\linewidth}{%
\textsuperscript{*}:~Work done during internship at Horizon Robotics. \\
\noindent\href{mailto:wei.ke@mail.xjtu.edu.cn}{\textsuperscript{†}:}~Corresponding author.
}}
\end{abstract}

\vspace*{-1.0em}
\begin{figure*}[t!]
  \centering
  \includegraphics[width=1\linewidth]{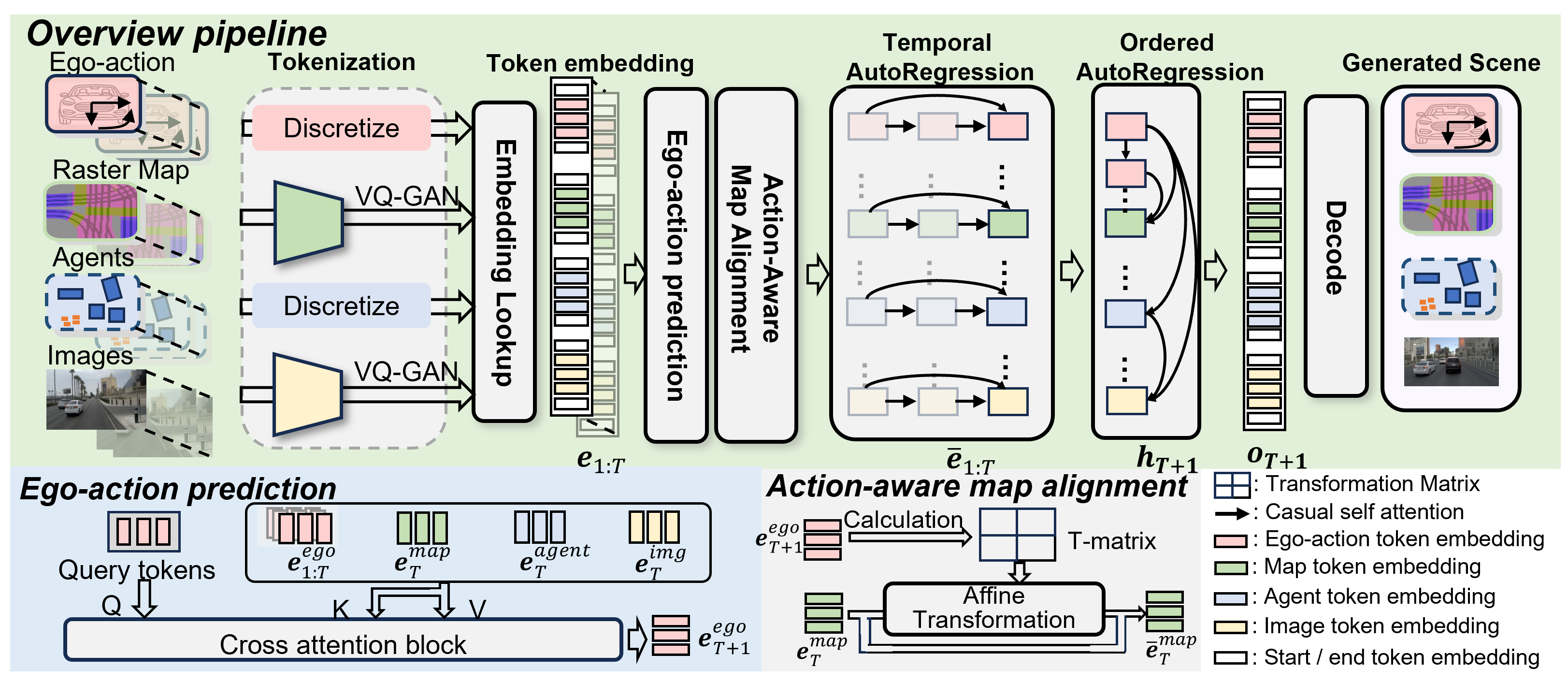}
    \vspace*{-2em}
  \caption{{\bf Pipeline of our \Methodname.} Given $T$ past frames of multimodal driving scenes, including ego-action, map, traffic agents, and image in each scene, each modality is tokenized into discrete tokens. The token embeddings are then processed through the Ego-action Prediction module, which forecasts the ego-action for $T+1$ time step. Using this predicted ego-action, the AMA module adjusts the map features. Next, the TAR module aggregates temporal information across sequences, while the OAR module ensures sequential modality prediction within each frame by autoregressively generating each token conditioned on the aggregated history information. Finally, the predicted tokens are fed to the decoder to obtain the next scene.}
  \vspace*{-1.5em}
  \label{fig:method}
\end{figure*}
\vspace{-1em}

\section{Introduction}
\label{sec:intro}
\vspace{-0.2cm}

Generative models are becoming increasingly essential across various domains~\cite{gen_fun, gen_GPT4, gen_img, gen_text2img}. In Autonomous Driving (AD) systems, generative models are utilized for producing diverse driving scenes—particularly those that are rare or not well-represented in datasets~\cite{agent_simu_magicdrive, agent_simu_drivedreamerv1, agent_simu_drivedreamerv2} or real-life. 
More importantly, generative models can be used to build a closed-loop simulation system that generates realistic interactive scenes to test AD systems without causing potential accidents before deployment~\cite{yang2023unisim, yang2024drivearena}. 
Such capabilities enhance the safety, adaptability, and reliability of autonomous driving~\cite{agent_simu_metadrive, AD_sparsedrive, AD_HEdrive, AD_UniAD} systems when dealing with the complexities of real-world driving conditions.

To generate comprehensive driving scenes, as described in Tab.~\ref{table:compare_mutilmoda}, most existing studies~\cite{img_gaia, Sovling_Mtion} focus primarily on generating two modalities. For instance, GUMP~\cite{Sovling_Mtion} and TrafficGen~\cite{gen_trafficgen} generate ego-action and agent motion within a given static map segment;
DriveDreamer~\cite{agent_simu_drivedreamerv1} and GAIA-1~\cite{img_gaia} are capable of generating images given initial conditions. While these approaches offer valuable insights, they face some limitations. The lack of map progression in GUMP and TrafficGen limits their realism, as real-world driving scenes involve dynamically evolving maps from the ego-vehicle’s viewpoint. DriveDreamer and GAIA-1 fail to forecast traffic agent motion, which limits control over agent behaviors and, consequently, constrains the generation of user-specific scenes.

In this work, we propose a {\textbf U}nified {\textbf M}ultimodal driving scene \textbf{Gen}eration framework (UMGen) for driving scenes, each containing four key modalities. Our framework integrates map prediction to enhance scene representation, enabling finer control over ego-action and agent behaviors.
Generating these modalities scene-by-scene while ensuring consistency presents a significant challenge. To address the diverse domains of each modality, we frame scene generation as a sequence scene prediction task. However, directly applying an AutoRegressive (AR) model is impractical due to the high token count from the four modalities and the length of the video data. To address this, we decompose the sequence prediction task into two stages: inter-frame prediction and intra-frame prediction. For inter-frame prediction, we propose the Temporal AutoRegressive (TAR) module, which models temporal evolution across frames using causal attention, allowing each token to be influenced only by its past states. 
For intra-scene prediction, we introduce the Ordered AutoRegressive (OAR) module, which captures relationships within each scene by enforcing a structured modality order, ensuring alignment and coherence across modalities. Together, TAR and OAR effectively capture cross-modal dependencies over time, enhancing alignment while reducing computational complexity.
To further maintain consistency between ego-action and map data, we introduce an Action-aware Map Alignment (AMA) module. AMA applies an affine transformation, based on predicted ego-action, to prior map features, aligning them with the ego-vehicle's movement. This re-alignment preserves coherence between ego-action and map modalities.

With these components, our model demonstrates strong capabilities in multimodal driving scene generation, as shown in Fig.~\ref{fig:Teaser_Big}, producing scene sequences that can span up to 60 seconds in duration. 
Additionally, thanks to the modality alignment capability provided by the AMA and OAR modules, our model can create specific multimodal-consistent scenes, like a cut-in situation, by controlling ego-vehicle and agents' actions.

\begin{table}[]
\tabcolsep=0.15cm
\begin{tabular}{l|cccc}
\hline
\multicolumn{1}{c|}{\multirow{2}{*}{Methods}} & \multicolumn{4}{c}{Modalities}                                                                                     \\
\multicolumn{1}{c|}{}                         & \multicolumn{1}{l}{Ego-action} & \multicolumn{1}{l}{Map} & \multicolumn{1}{l}{Agent} & \multicolumn{1}{l}{Image} \\ \hline
Drivedreamer~\cite{agent_simu_drivedreamerv1}                                   & \cmark                           & \xmark
                      & \xmark
                         & \cmark                            \\
TrafficGen~\cite{gen_trafficgen}                                    & \cmark                               &\xmark
                       & \cmark                             & \xmark
                        \\ 
GAIA-1~\cite{img_gaia}
& \cmark                           & \xmark
                       & \xmark
                  & \cmark                              \\ 
GUMP~\cite{Sovling_Mtion}                                          & \cmark                                & \xmark
                & \cmark                       & \xmark
                        \\ \hline
\Methodname~(Ours)                                   & \cmark                                & \cmark                         & \cmark                            & \cmark

\end{tabular}
\vspace*{-1em}
\caption{Comparison of different driving scene generation methods across four modalities. 
}
\vspace*{-2em}

\label{table:compare_mutilmoda}
\end{table}

In summary, \Methodname~offers several key contributions:
\begin{itemize}

\item We introduce a novel generative framework that integrates four distinct modalities—ego-action, road users, traffic maps, and images—with the flexibility to incorporate additional modalities, enhancing the scene representation and fidelity of driving scene generation.

\item We design a computationally efficient AR approach, comprising TAR and OAR modules, to capture both inter-frame and intra-frame dependencies, enabling realistic scene generation with reduced computational costs.

\item We introduce an AMA module to apply affine transformations to map features with the ego-vehicle's movement， ensuring consistency between ego-action and map. 
\end{itemize}

Our experimental results provide both quantitative and qualitative evidence of our method's effectiveness, showing that \Methodname~enables user-defined, ego-centric scenario generation adaptable to specific driving conditions.

~~~~~
\vspace{-2.5em}
\section{Related Work}
\label{sec:Related_Work}
\vspace{-0.5em}
{\bf Autoregressive generative models.}
In recent years, AR generative models have achieved significant success in natural language processing and image generation, as evidenced by models such as GPT-2~\cite{other_GPT2} and VQGAN~\cite{other_VQGAN}. This architecture has since been adapted for autonomous driving scene generation, with methods like~\cite{agent_simu_trajeglish, simu_agent_motionlm, agent_simu_behaviorGPT} leveraging it to predict agent behaviors, and GAIA-1~\cite{img_gaia} applying it to generate high-quality driving videos sequentially.
Our model, however, extends beyond generating limited-modality data by targeting the generation of multimodal data that collectively compose driving scenes. This extension results in a substantial increase in token count, especially as the number of scene frames grows, making the vanilla AR~\cite{img_gaia} approach—where tokens from different scene frames are concatenated into a single sequence—computationally prohibitive. 
Additionally, applying the AR mechanism solely along the temporal dimension—without incorporating an intra-frame AR mechanism—may lead to inconsistencies across modalities within the same frame~\cite{agent_simu_behaviorGPT, simu_agent_motionlm}.
By contrast, our method consists of parallel inter-frame temporal prediction with intra-frame AR decoding which balances efficiency and consistency to enable multimodal driving scene generation.

{\bf Driving scene sequence generation.}
Traditional methods rely on hand-crafted rules and human prior knowledge~\cite{agent_simu_metadrive, agent_simu_nuplan, re_scene_gen_manmade_gen0, re_scene_gen_manmade_gen1} to generate driving scene sequences, limiting their ability to capture the diversity and realism of the real world. 
Recently, data-driven deep learning approaches have gained traction, but most focus on generating a limited set of modalities. For example, models like~\cite{agent_simu_symphony, gen_unigen, agent_simu_behaviorGPT,gen_lctgen} generate diverse agent trajectories based on map segments but fail to handle the emergence and disappearance of agents. GUMP~\cite{Sovling_Mtion} addresses this challenge through a GPT-like~\cite{other_GPT2} generative framework. However, the lack of map generation confines trajectories to specific map segments with static perspectives, which differs from the real world, where vehicles are observed from the dynamic viewpoint of the ego-vehicle. While DriveDreamer2~\cite{agent_simu_drivedreamerv2} addresses this limitation by generating maps, it employs two separate networks for map and video generation, resulting in the loss of explicit per-frame control. Other models like Dreamforge~\cite{agent_simu_magicdrive, img_gaia, agent_simu_drivedreamerv1, img_dreamforge} produce high-quality driving videos based on predefined traffic dynamics, which restricts interactivity and user-defined scene sequence generation. 
In contrast, UMGen generates four essential modalities—ego-action, map, agent, and image—enriching scene representation within a unified framework and providing finer control over user-specific scene sequence generation.
\vspace{-0.2cm}
\section{Method}
\label{sec:method}
\vspace{-0.15cm}
\subsection{Problem Setup}
\label{sec:sub:form}
\vspace{-0.15cm}
Our method generates multimodal driving scenes in an autoregressive scene-by-scene manner, starting from either a provided or self-generated initial scene sequence. To achieve this, we tokenize each scene element into sequential tokens, transforming the generation process into a scene token prediction task. These elements include the ego-vehicle’s action, a traffic map, other scene agents, and the camera image.
Specifically, the ego-vehicle action \( \mathbf{{a}}^{AV}_t \) represents displacements along the $x$ and $y$ axes, as well as angular changes relative to the previous timestamp. The traffic map \( \mathbf{{m}}_t \) depicts road layouts and situational details surrounding the ego vehicle, while the \( N_{a} \) agents \( \{\mathbf{{a}}^{(i)}_t\}_{i=1}^{N_{a}} \) represent each agent's coordinates, size, vehicle, head and category. This combination of data provides a comprehensive view of the scene's dynamic and static elements. Finally, the camera image \( \mathbf{{v}}_t \) adds visual context. Thus, the multimodal scene at time \( t \) is defined as \( {s}_t = \{ \mathbf{{a}}^{AV}_t, \mathbf{m}_t, \{\mathbf{{a}}^{(i)}_t\}_{i=1}^{N_{a}}, \mathbf{{v}}_t \} \). Therefore, our task is to learn a neural network to generate the next scene ${s}_{t+1}$, conditioned on ${s}_{1:t}$.

\subsection{Framework}
\label{sec:sub:net}
\vspace{-0.15cm}
Throughout the following formulation, we assume the availability of the past $T$ scenes as inputs to predict the scene at the \( T+1 \) timestep. The overall pipeline of our model is demonstrated in Fig.~\ref{fig:method}, which comprises several core components: \textit{ Tokenization}, an \textit{Ego-action prediction} module, an \textit{Action-aware Map Alignment (AMA)} module, a \textit{Temporal AutoRegressive (TAR)} module, and an \textit{Ordered Autogressive (OAR)} module. We discuss each component below.

\textbf{Tokenization.}\label{sec:tokenlize }~
The \textit{ego-action} and \textit{agent} attributes are tokenized through discretization, whereas the \textit{raster map} and \textit{image} data are tokenized using pretrained VQ-GAN models~\cite{other_VQGAN}.
Discrete tokens from modalities are arranged in a fixed sequence within each frame: ego-action, map, agent, and image.  This sequence reflects the causal flow: the ego vehicle's actions modify the observable traffic map and influence the behavior of surrounding agents. These interactions are ultimately captured in the camera view.
Using this order and the tokenizer functions, we obtain the ordered token at $t-$th scene as follows:
\vspace{-0.2cm}
\begin{equation}
\vspace{-0.2cm}
\textbf{z}_t = g\left( {s}_t \right)  = \left[ \mathbf{z}_t^{\text{ego}}, \mathbf{z}_t^{\text{map}}, \mathbf{z}_t^{\text{agent}}, \mathbf{z}_t^{\text{image}} \right],
\vspace{-0.2cm}
\end{equation}
where $\mathbf{z}_t\in \mathbb{N}^{N}$, $N$ denotes the total number of tokens obtained by concatenating tokens from all modalities and $g$ represents the tokenizer function. Unlike vanilla AR approaches~\cite{img_gaia} that concatenate tokens from multiple frames into a single sequence, we maintain tokens in a structured per-frame format. Consequently, the tokenized scene sequence across \( T \) timesteps is organized as:
$\mathbf{Z} = \left[ \mathbf{z}_1, \mathbf{z}_2, \dots, \mathbf{z}_{T} \right], \mathbf{Z}  \in \mathbb{N}^{T \times N}, $ where tokens of each frame retain their intra-frame order. 
This structured token arrangement enables parallel aggregation of temporal information, forming the basis for subsequent modules. To ensure a fixed token count per frame, we apply padding or jointly sample agents across $T$ frames, preserving a consistent set of objects across all frames for temporal coherence.

\textbf{Token Embedding.}~
Each token is embedded into latent features through learnable token embedding codebooks. All embeddings are then summed with a learnable positional embedding and projected to a unified dimension using a Multi-Layer Perceptron (MLP) head:
\vspace{-0.2cm}
\begin{equation}
\textbf{e}_{t}^{i} = \text{MLP} \left( \textit{Embed}(\mathbf{z}_{t}^{i}) + \textit{PE}(i) \right),  \textbf{e}_{t}^{i} \in \mathbb{R}^{D}
\vspace{-0.2cm}
\end{equation} 
where $i$ is the position index in the token sequence of $\mathbf{z_t}$, \textit{PE} is the learnable position embedding and \textit{Embed} is the learnable token embedding. 
Thus, the token sequence representing an entire scene can be expressed as: 
\begin{equation}
\begin{aligned}
\mathbf{e}_t = & \underbrace{\big[\mathbf{e}_t^1, \dots, \mathbf{e}_t^{n_{ego}}}_{\mathbf{e}_t^{ego}}, \, \underbrace{\mathbf{e}_t^{n_{ego}+1}, \dots, \mathbf{e}_t^{n_m}}_{\mathbf{e}^{map}_t}, \, \\
&~~~~~~~~~~~~~~~ \underbrace{\mathbf{e}_t^{n_m+1}, \dots, \mathbf{e}_t^{n_a}}_{\{\mathbf{e}^{agent}_{t}\}}, \, \underbrace{\mathbf{e}_t^{n_a+1}, \dots, \mathbf{e}_t^{n_v}\big]}_{\mathbf{e}^{img}_t},
\end{aligned}
\vspace{-0.2cm}
\end{equation}
where $\mathbf{e}_t \in \mathbb{R}^{N \times D} $, and the indices \( n_{ego} \), \( n_m \), \( n_a \), and \( n_v \) correspond to the cumulative token counts for each modality. The terms \( \mathbf{e}_t^{ego} \), \( \mathbf{e}_t^{map} \), \( \mathbf{e}_t^{agent} \), and \( \mathbf{e}_t^{img} \) represent the tokens derived from each modality at time step \( t \).
More details about tokenization and token embedding can be found in the supplementary material section B.


\textbf{Ego-action Prediction.} We first predict the next ego-action \(\mathbf{e}^{ego}_{T+1}\), using it as a prior for generating the remaining modalities. Historical ego actions are aggregated via a cross-attention mechanism to capture the ego vehicle's intentions, which is then integrated with the current environment state \(\mathbf{E}_{T} = \left[ \mathbf{e}^{map}_{T},\mathbf{e}^{agent}_{T}, \mathbf{e}^{img}_{T} \right]\) through another cross-attention mechanism:  
\vspace{-0.2cm}
\begin{equation}
\vspace{-0.15cm}
\begin{aligned}
\mathbf{u}_{T} &= \textit{CA}_{\text{hist}} \left( \mathbf{Q} = \mathbf{q}, \mathbf{K} = \{\mathbf{e}^{ego}_{1:T}\}, \mathbf{V} = \{\mathbf{e}^{ego}_{1:T}\} \right), \\
\mathbf{e}^{ego}_{T+1} &= \textit{CA}_{\text{env}} \left( \mathbf{Q} = \mathbf{u}_{T}, \mathbf{K} = \{\mathbf{E}_{T}\}, \mathbf{V} = \{\mathbf{E}_{T}\} \right),
\end{aligned}
\end{equation}
where \(\mathbf{q}\) represents the query tokens, and \(CA_{\text{hist}}\) and \(CA_{\text{env}}\) denote the two cross-attention blocks. This module serves as a basic planning component.

\textbf{Action-aware Map Alignment (AMA). }~
The AMA module is motivated by the continuity of map features across adjacent frames: adjusting current map features based on the ego-vehicle’s displacement offers a strong prior for the next timestep.
As a result, we first rearrange the map feature vector $\mathbf{e}^{map}_T$ to obtain a spatial map representation, denoted as $\mathbf{F}_{T} = \Gamma(\mathbf{e}^{map})$, where $\Gamma: \mathbb{R}^{N_{m} \times D} \rightarrow \mathbb{R}^{H \times W \times D}$. Here, $N_{m}$ refers to the number of tokens representing the map, and $H$ and $W$ denote the height and width of the resulting spatial map, respectively.
An affine transformation, parameterized by \(\theta\), \(dx\), and \(dy\) (derived  from \(\mathbf{e}^{AV}_{T+1}\)), generates a sampling grid \( \mathbf{G} \) that maps coordinates from the original map representation \( \mathbf{F}_{T} \) to the transformed locations. For each point \((x, y)\) in \( \mathbf{F}_{T} \), the transformed coordinates \((x', y')\) are given by:
\vspace{-0.25cm}
\begin{equation}
\vspace{-0.15cm}
\begin{bmatrix} x' \\ y'  \end{bmatrix} = \mathbf{G}(x, y) =
\begin{bmatrix}
\cos \theta & -\sin \theta \\
\sin \theta & \cos \theta 
\end{bmatrix} 
\begin{bmatrix} x \\ y \end{bmatrix} + 
\begin{bmatrix} dx \\ dy \end{bmatrix}.
\end{equation}
Using the sampling grid \( \mathbf{G}(x, y) \), we interpolate \( \mathbf{F}_{T} \), producing the transformed map representation 
\( \mathbf{\bar{F}}_{T} \in \mathbb{R}^{H \times W \times D} \) that represents the transformed map features.
Finally, we flatten the representation $\bar{\mathbf{F}}_{T}$ and add it with $\mathbf{e}^{map}_{T} $ to yield transformed map representation 
$\bar{\mathbf{e}}^{map}_{T}$:\vspace{-0.1cm}
\begin{equation}
\bar{\mathbf{e}}^{map}_{T} = \text{Flatten}(\bar{\mathbf{F}}_{T}) + {\mathbf{e}}^{map}_{T}.
\end{equation}
~~~\textbf{Inter-frame Temporal Autoregression (TAR).} With the predicted ego-action and transformed map features, the scene state is updated as \( \mathbf{\bar{e}}_{T} = [\ \mathbf{e}^{ego}_{T+1}, \mathbf{\bar{e}}^{map}_{T}, \mathbf{e}^{agent}_{T}, \mathbf{e}^{img}_{T} ]\). For previous frames, we similarly obtain the ego-action and apply the AMA module to map features, resulting in \( \mathbf{\bar{e}}_{1:T} \in \mathbb{R}^{T \times N \times D} \). 
A causal self-attention (CSA) block is applied to \( \overline{\mathbf{e}}_{1:T} \), where each token attends only to tokens at the same position across previous scenes:
\vspace{-0.1cm}
\begin{equation}
\vspace{-0.15cm}
\bar{\mathbf{e}}_{T+1}^i = \text{CSA}\left( \bar{\mathbf{e}}_{1}^i, \bar{\mathbf{e}}_{2}^i, \dots, \bar{\mathbf{e}}_{T}^i \right),  \bar{\mathbf{e}}_{T+1}^i \in \mathbb{R}^{D}.
\end{equation}
Where \( i \) denotes the token's position. This design captures temporal dependencies efficiently by processing tokens in parallel, thereby avoiding the computational costs associated with performing attention mechanism on the long sequence generated by token concatenation~\cite{img_gaia}.
Subsequently, a bidirectional self-attention (SA) block facilitates information exchange between tokens within the same scene, yielding a coarse prediction $\mathbf{h}_{T+1}$ for the next scene:
\vspace{-0.5cm}  
\begin{equation}
\mathbf{h}_{T+1} = \text{SA} \left( 
\mathbf{\bar{e}}_{T+1}^1, \dots, \mathbf{\bar{e}}_{T+1}^N
\right), \mathbf{{h}}_{T+1} \in \mathbb{R}^{N \times D}.
\vspace{-0.1cm}  
\end{equation}
~~~~\textbf{Intra-frame Ordered Autoregression (OAR).}~ 
The OAR module leverages TAR predictions as temporal priors to enhance inter-frame continuity while maintaining intra-frame consistency through GPT-like autoregressive decoding. To predict the $i$ th token embedding \(\mathbf{o}^{i}_{T+1}\), the OAR combines the previous token embedding \(\mathbf{o}^{1:i-1}_{T+1}\) with the TAR-derived feature \(\mathbf{h}_{T+1}\), incorporating temporal priors at each prediction step.
The OAR then applies a causal self-attention block to sequentially predict each token embedding \(\mathbf{o}^{i}_{T+1}\). Following this, a projection head is applied to \(\mathbf{o}_{T+1}^i\), generating a probability vector \( \mathbf{p}^{(OAR,i)}_{T+1} \in \mathbb{R}^{K} \), where \(K\) represents the codebook size:
\vspace{-0.1cm}
\begin{gather}
w_j = {\mathbf{o}}_{T+1}^j + \mathbf{h}_{T+1}^{j+1}, \quad j = 1, \dots, i-1, \\
{\mathbf{o}}_{T+1}^{i} = \text{CSA} \left( w_1, w_2, \dots, w_{i-1} \right), \\
\mathbf{p}^{(OAR, i)}_{T+1} = \text{Softmax}(\text{MLP}(\mathbf{o}_{T+1}^i)).
\vspace{-4cm}
\end{gather}
~~~~~A Top-k sampling strategy~\cite{other_GPT2} is then used on \( \mathbf{p}^{(OAR, i)}_{T+1} \) to select token \(\mathbf{z}^{i}_{T+1}\), which is transformed into embedding for the next prediction. This autoregressive structure aligns tokens contextually, preventing conflicts and enhancing modality coherence.

\textbf{Loss functions.}
Additional to the OAR predicted token probability $\mathbf{p}^{OAR}_{T+1}$, we also apply a projection head to $\mathbf{h}_{T+1}$ to obtain  $\mathbf{p}^{TAR}_{T+1}$. We then apply cross-entropy loss functions to calculate the total loss:
\vspace{-0.2cm}
\begin{gather}
\mathbf{p}^{(TAR)}_{T+1} = \text{Softmax}(\text{MLP}(\mathbf{h}_{T+1})), \\
\mathcal{L}_{\text{total}} = CE(\mathbf{p}^{OAR}_{T+1}, \mathbf{z}_{T+1}) + CE(\mathbf{p}^{TAR}_{T+1}, \mathbf{z}_{T+1}),
\vspace{-2cm}
\end{gather}
\vspace{-0.1cm}
where $CE$ represents the cross-entropy loss function.

\begin{figure*}[t]
  \centering
  \includegraphics[width=1\linewidth]{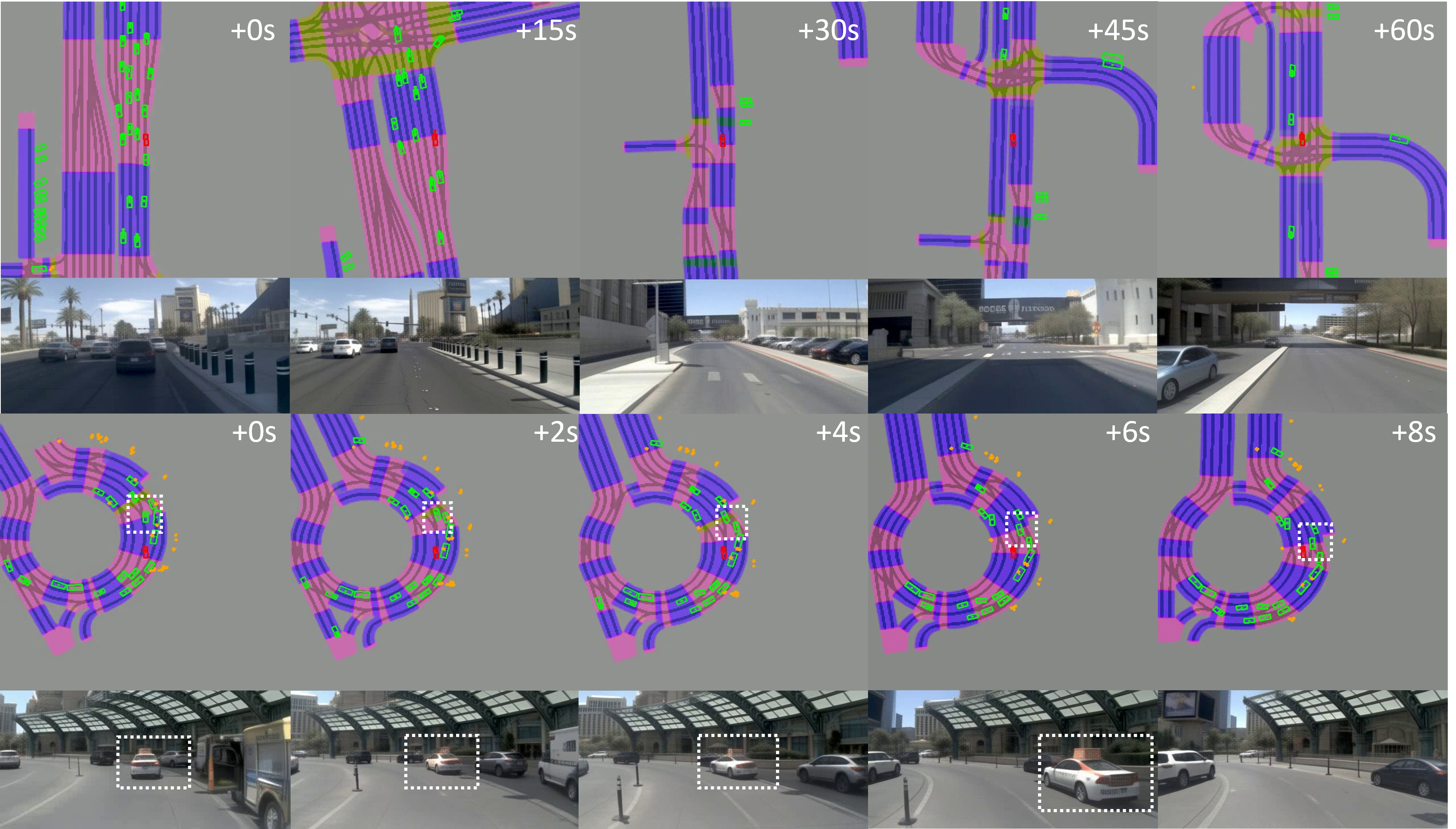}
  \vspace{-0.6cm}
  \caption{{\bf Generated multimodal driving scenes by \Methodname}: The generated scenes evolve continuously from the ego vehicle's perspective.  Red Box: ego-vehicle, Green Box: cars, Orange Box: pedestrians or cyclists, Arrow: agent velocities.}
  \vspace{-0.5cm}
  \label{fig:dreamvideos}
\end{figure*}

\section{Experiments}
\label{sec:experiments}
Our experiments evaluate the effectiveness of~\Methodname~in both multimodal driving scene sequence generation and user-guided specific scene sequence generation. We also demonstrate that our model can generate realistic initial scenes, serving as the starting point for sequence generation. Finally, we conduct ablation studies to confirm the critical role of each module.
\vspace{-0.15cm}
\subsection{Experimental Settings}
\label{subsec:experiment_setting}
{\bf Training details.}  
We randomly select a 21-frame sequence at each iteration, allowing the model to leverage up to 20 past frames. The training process is conducted over 300 epochs on 32 RTX4090 GPUs for two days.

{\bf Evaluation details.}
Our experiments leverage two public datasets: nuPlan~\cite{agent_simu_nuplan} and the Waymo Open Motion Dataset (WOMD)~\cite{other_dataset_waymo}. We assess the realism of the generated initial scenes using Maximum Mean Discrepancy (MMD)~\cite{gen_trafficgen, gen_lctgen, gen_unigen} scores, following the experimental setup in TrafficGen~\cite{gen_trafficgen} on both nuPlan and WOMD datasets. The MMD metric quantifies the distributional divergence between the generated and ground-truth agent attributes. When generating initial scenes, we temporarily discard the TAR module since there is no temporal information needed.
We present visualizations of the generated multimodal driving scene sequences based on the provided initial frames from nuPlan, verifying both the generation of diverse multimodal scenes and user-specified scene segments. 
For assessing the computational efficiency of the TAR module relative to the vanilla AR approach, we measure the per-token inference time and peak GPU memory usage during inference. 
To verify the efficacy of the OAR module in maintaining token consistency, we use both the MMD metric and the average collision rate (CR) among generated road users. Lower CR values indicate reduced conflicts between tokens, demonstrating better consistency. Further training and evaluation details are provided in the appendix.

\vspace{-0.15cm}
\subsection{Driving Scene Sequence Generation}
\vspace{-0.15cm}
\label{subsec:extrapolation}
Unlike other methods that rely on supplementary conditions and generate limited modalities~\cite{Sovling_Mtion, agent_simu_drivedreamerv1, agent_simu_magicdrive}, our model is capable of generating diverse multimodal scene sequences from only given initial frames. To demonstrate the dynamic evolution of our multimodal scenes, as well as the temporal and multimodal consistency of the generated scenes, we present two video samples at different time intervals.

As shown in the first row of Fig~\ref{fig:dreamvideos}, we display the generated scenes at 15-second intervals, showing how all modalities evolve as the ego-vehicle moves. 
Specifically, the ego vehicle performs maneuvers such as lane changes, turns, and straight driving. The map modality generates a variety of road elements, including intersections, curves, and straight paths according to the ego vehicle's movement. The number of agents decreases from numerous at the beginning to fewer on narrower roads. Furthermore, the image modality presents the corresponding visualization that reflects these dynamics. This capability of on-the-fly multimodal generation allows our approach to simulate a broader spectrum of driving worlds. The second row of Fig~\ref{fig:dreamvideos} shows 2-second intervals of the ego vehicle passing a hotel entrance. Dashed boxes are used to mark the same car across these frames, showing that its position and movement on the map remain consistent with those in the generated images. This multimodal consistency and temporal continuity provide a solid foundation for generating specified multimodal scenes in the subsequent sections.

\vspace{-0.15cm}
\subsection{Initial Scene Generation}
\label{subsec:Scene_gen}
\vspace{-0.15cm}
The ability to generate realistic initial scenes is paramount for facilitating the generation of driving scene sequences. To evaluate the generated initial scenes, we follow the experimental setting of TrafficGen~\cite{gen_trafficgen} and evaluate the MMD score on the WOMD~\cite{other_dataset_waymo} and nuPlan~\cite{agent_simu_nuplan}.
As shown in Table~\ref{table:MMD_Waymo} and Table \ref{table:nuplan}, our~\Methodname~achieved lower MMD values compared to other approaches, indicating that it generates more realistic scenes. 

\begin{table}[]
\begin{tabular}{c|cccc}
\hline
\multicolumn{1}{c|}{Method} & \multicolumn{1}{c}{Position} & \multicolumn{1}{c}{Heading} & \multicolumn{1}{c}{Size} & \multicolumn{1}{c}{Velocity} \\ \hline
TrafficGen\cite{gen_trafficgen}                 & 0.1451                       & 0.1325                      & 0.0926                   & 0.1733                       \\
LCTGen\cite{gen_lctgen}                      & 0.1319                       & 0.1418                      & 0.1092                   & 0.1938                       \\
SceneGen\cite{gen_scenegen}                    & 0.1362                       & 0.1307                      & 0.1190                   & 0.1772                       \\
UniGen\cite{gen_unigen}                      & 0.1208                       & 0.1104                      & 0.0815                   & 0.1591                       \\ 
Ours                        & \textbf{0.0730}              & \textbf{0.0550}             & \textbf{0.0533}          & \textbf{0.1502}                       \\ \hline
\end{tabular}
\vspace{-0.3cm}
\caption{{\bf MMD results on Waymo Open Motion Dataset}}
\label{table:MMD_Waymo}
\vspace{-0.3cm} 
\end{table}

\begin{table}[]
\begin{tabular}{c|cccc}
\hline
Method     & Position        & Heading         & Size            & Velocity        \\ \hline
TrafficGen* & 0.0926          & 0.0799          & 0.0856          & 0.0988          \\
Ours       & \textbf{0.0828} & \textbf{0.0682} & \textbf{0.0674} & \textbf{0.0760} \\ \hline
\end{tabular}
\vspace{-0.3cm}
\caption{{\bf MMD results on nuPlan.} {$\star$}: Results reproduced by us under the same experimental setting }
 \label{table:nuplan}
\vspace{-0.7cm}

\end{table}

\begin{figure*}[]
  \centering
  \includegraphics[width=1\linewidth]{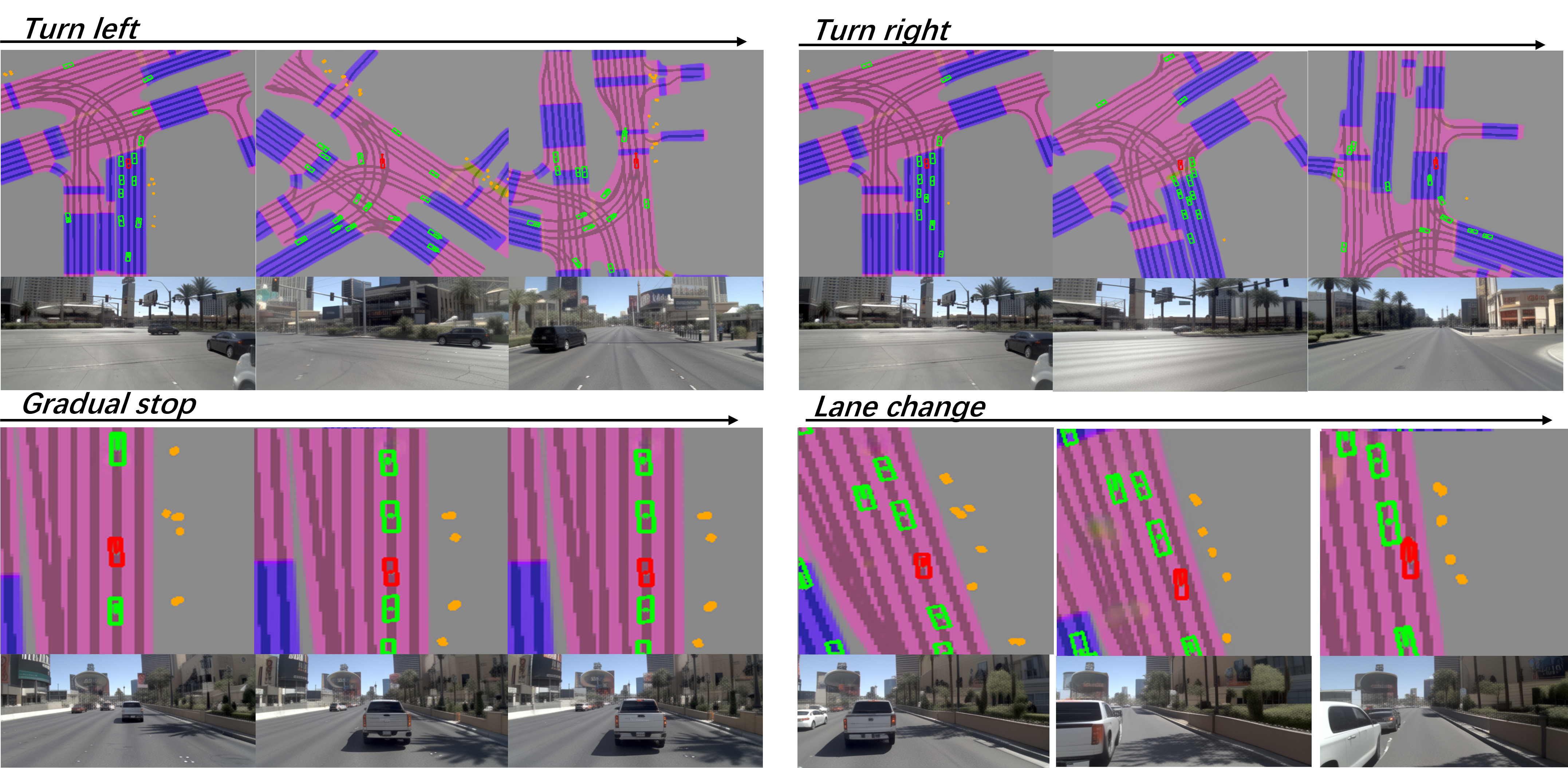}
   \vspace*{-2em}
  \caption{{\bf Generated scenes with input ego actions}. The first row shows the interactive control of the ego vehicle to perform left and right turns. The second row shows the ego vehicle, initialized with a right-turn velocity or a gradual deceleration to a stop. Red box: ego-vehicle, green box: vehicle, orange box: pedestrians or cyclists, arrow: agent velocity.
  }
  \vspace*{-1em}
  
  \label{fig:drive_everywhere}
  \vspace*{-0.5em}

\end{figure*}

\vspace*{-0.5em}
\subsection{User-Guided Scene Sequence Generation}
\vspace*{-0.5em}

\label{subsec:user_scene}
The ability to generate interactive scene sequences that replicate real-world conditions, along with control over the ego vehicle's movements within these scenes, is essential for validating AD systems. This section, therefore, focuses on user-guided scene generation, highlighting our model's flexibility in adapting to diverse ego-vehicle controls and simulating various traffic interactions.

{\bf Interactive ego-vehicle control.} 
To illustrate the model's ability to control ego-vehicle actions, we present visualization from two distinct generated scene sequences.
As shown in the first row of Fig.\ref{fig:drive_everywhere}, we actively control the ego-vehicle to execute either a left turn or a right turn. The map undergoes corresponding rotations and transformations based on the ego vehicle's actions, while the image modality provides a visualization of these changes. Notably, although the scene observed after the right turn does not exist in the dataset, our model generates a corresponding multimodal scene. 
In another scene (second row), instead of following the dataset's "gradual stop" action (left), we assign the ego vehicle a one-frame right-turn velocity. The model autonomously executes maneuvers of lane change and overtaking (right), showcasing its ability to interpret intention, predict maneuvers, and update modalities. Leveraging multimodal generation, we observe the ego vehicle avoids collision with the car ahead despite close proximity—an insight that would be challenging to achieve with models restricted to image-only predictions.

\begin{figure}[]
  \centering
  \includegraphics[width=1\linewidth]{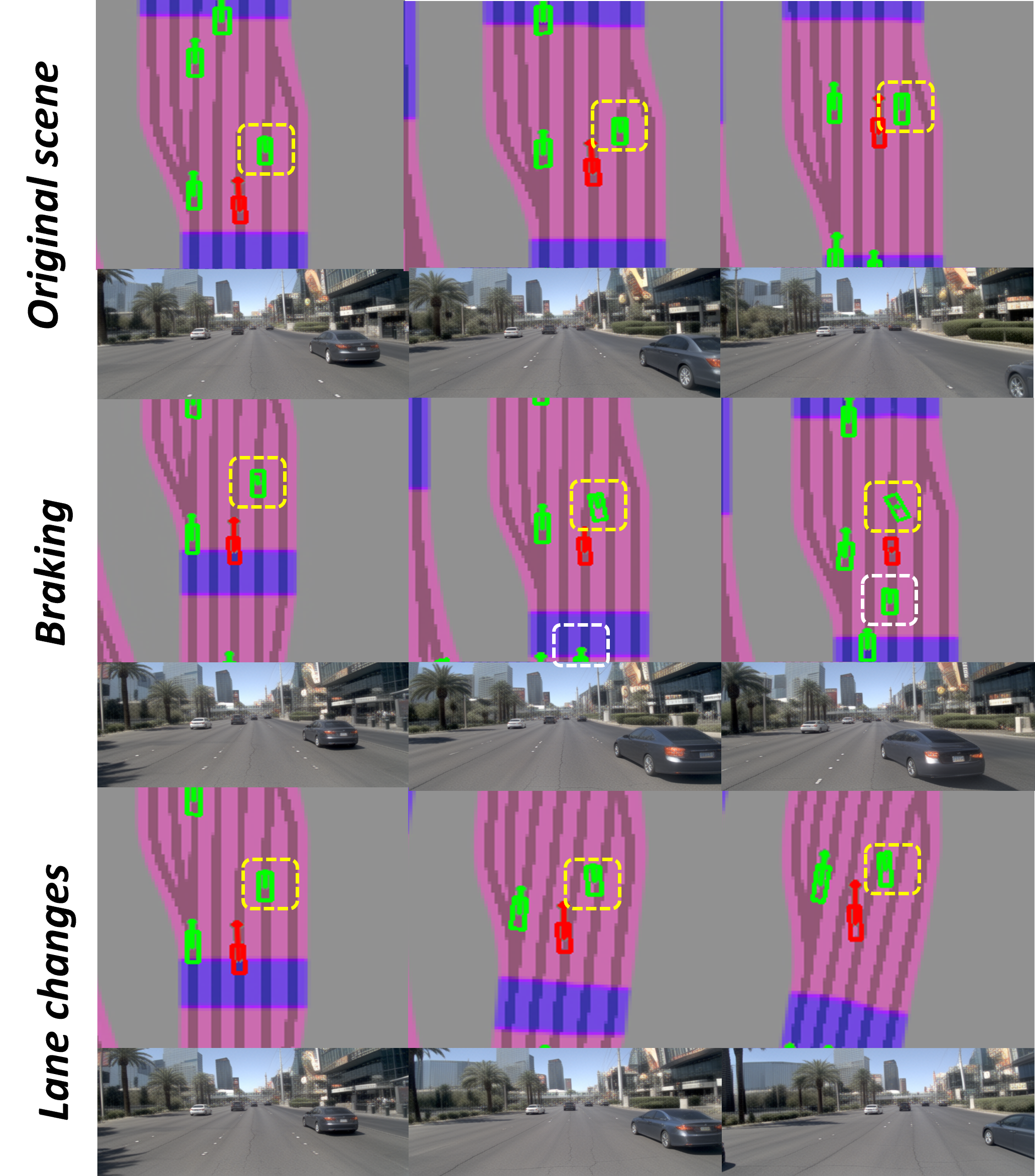}
  \vspace*{-2em}

  \caption{{\bf Customized scenario generation by \Methodname}: The first row presents the original scene from the dataset.  We assign a forward-left velocity to the vehicle highlighted by the yellow dashed line box and the ego-vehicle spontaneously takes a braking action (second row). Alternatively, we can actively control the ego-vehicle to perform a lane change (third row). Red box: ego-vehicle, green box: vehicle, arrow: agent velocity.   }
 
  \label{fig:control_cut_in}
\end{figure}

{\bf User-specified agent control.}
\Methodname~also provides flexible control over other road users, enabling the generation of specific scenes. Moreover, it can simultaneously simulate how these controls impact the behavior of other road users.
For example, in the scene shown in Fig.\ref{fig:control_cut_in}, we simulate a sudden cut-in maneuver by assigning a forward-left velocity to the agent marked by yellow dashed boxes. In response, the ego-vehicle executes emergency braking (the velocity arrow quickly turns shorter), as shown in the second row, with a trailing agent (highlighted with white dashed boxes) also decelerating in reaction to the ego-vehicle’s braking. Compared to the ego's spontaneous braking action, we can also simultaneously control the ego-vehicle to perform a lane change to avoid the collision, as shown in the third row. This shows the flexible controllability of our model. Notably, such agent control through speed settings and realistic interaction simulation are two key features that simplify the creation of specific scenes. Speed control offers a human-like way to adjust agent behavior, while interaction simulation enables agents to respond dynamically to each other. Without these capabilities, generating such situations would require manual calculations of agent positions and interactions between road users.

\begin{figure}[]
  \vspace*{-1em}
  \centering
  \includegraphics[width=1\linewidth]{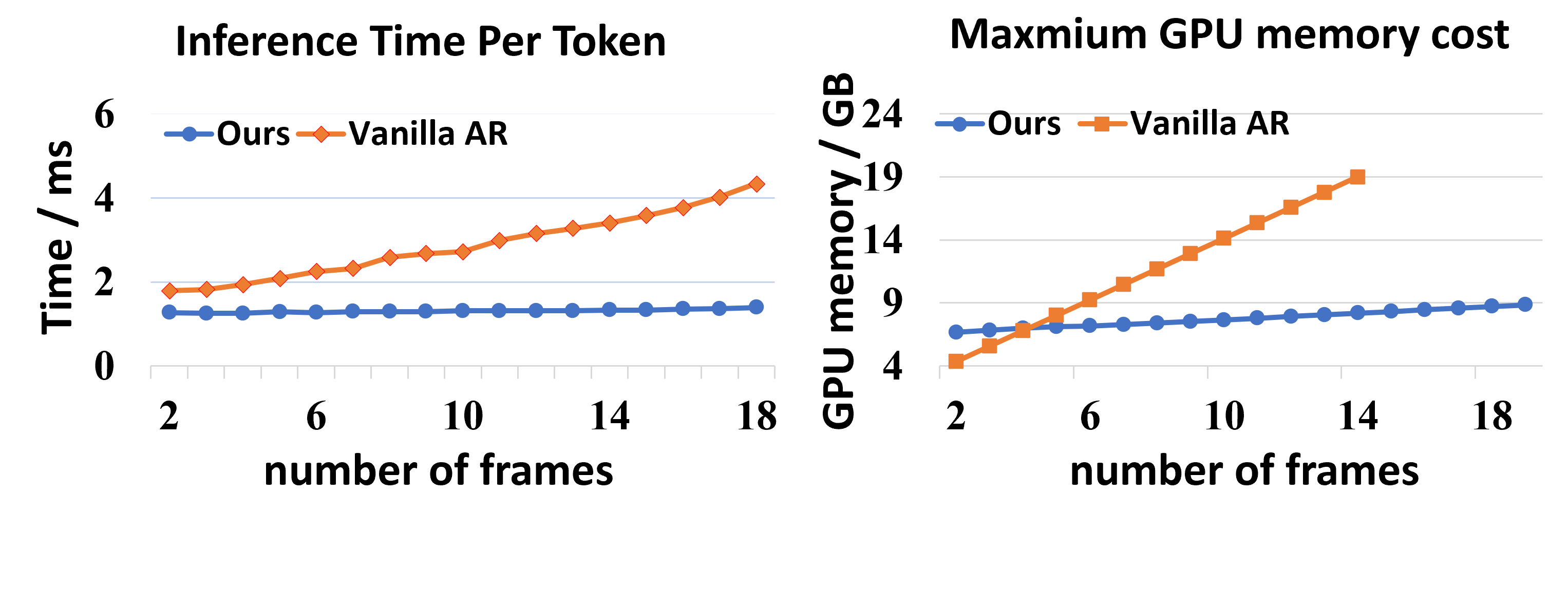}
  \vspace*{-2em}
  \caption{{\bf Comparison of peak GPU memory usage and per-token inference time} for the vanilla AR model and our \Methodname.}
  \vspace*{-2em}
  
  \label{fig:infer_time_and_GPU_Cost}
\end{figure}

\begin{table}[]
\begin{tabular}{l|l|llll}
\hline
\multicolumn{1}{c|}{\multirow{2}{*}{Method}} & \multicolumn{1}{c|}{\multirow{2}{*}{CR($\%$)}} & \multicolumn{4}{c}{MMD}                                                                                  \\
\multicolumn{1}{c|}{}                        & \multicolumn{1}{c|}{}                          & \multicolumn{1}{c}{Posi} & \multicolumn{1}{c}{Size} & \multicolumn{1}{c}{Head} & \multicolumn{1}{c}{Vel} \\ \hline
\Methodname-T                              & 5.6                               & 0.040                    & 0.043                    & 0.041                    & 0.049                   \\
\Methodname                                  & \textbf{5.1}                                  & \textbf{0.038}           & \textbf{0.037}           & \textbf{0.040}           & \textbf{0.036}          \\ \hline
\end{tabular}
\vspace*{-0.5em}
  \caption{{\bf MMD and collision rate results.} \Methodname-T represents \Methodname~without OAR module. CR: averaged agent collision rates. Vel: Velocity. }
\vspace*{-2em}
\label{table:compare_PAR}

\end{table}

\subsection{Ablation study}
\vspace*{-0.5em}
In this section, we analyze each component of \Methodname.

{\bf Effectiveness of TAR}. 
Our TAR module applies causal self-attention exclusively to tokens at the same position across historical frames, reducing the time complexity from \( O((TN)^2) \) in the vanilla AR model to \( O(T^2) \) for modeling temporal dependencies, where \(T\) is the number of frames and \(N\) is the number of tokens per frame. This design enhances computational efficiency and significantly reduces GPU memory usage.
We conducted comparative experiments to validate these benefits using the vanilla AR model and \Methodname, monitoring peak GPU memory usage and per-token inference time. As shown in Fig.~\ref{fig:infer_time_and_GPU_Cost}, \Methodname~maintains stable performance, in contrast to the vanilla AR model, whose GPU memory consumption and inference time increase sharply with the addition of frames, highlighting the challenges of long-horizon scene generation.

{\bf Decoding without OAR}. Our OAR module autoregressively decodes tokens within a frame aiming to prevent inconsistency between tokens. To analyze the impact of this component, we created a variant of \Methodname~by removing the OAR module, directly decoding from the TAR output. This modified version, referred to as \Methodname-T, was evaluated by generating scene sequences based on the initial frames provided by the nuPlan validation set.
Performance was measured using the average collision rates between agents and MMD scores on generated scene sequences. As shown in Table~\ref{table:compare_PAR}, \Methodname-T exhibits higher MMD values, indicating that its generated scenarios appear less realistic. More importantly, the elevated agent collision rates suggest that \Methodname-T struggles to effectively capture intra-frame token relationships, resulting in more frequent conflicts.

\begin{figure}[]
  \centering
  \includegraphics[width=1\linewidth]{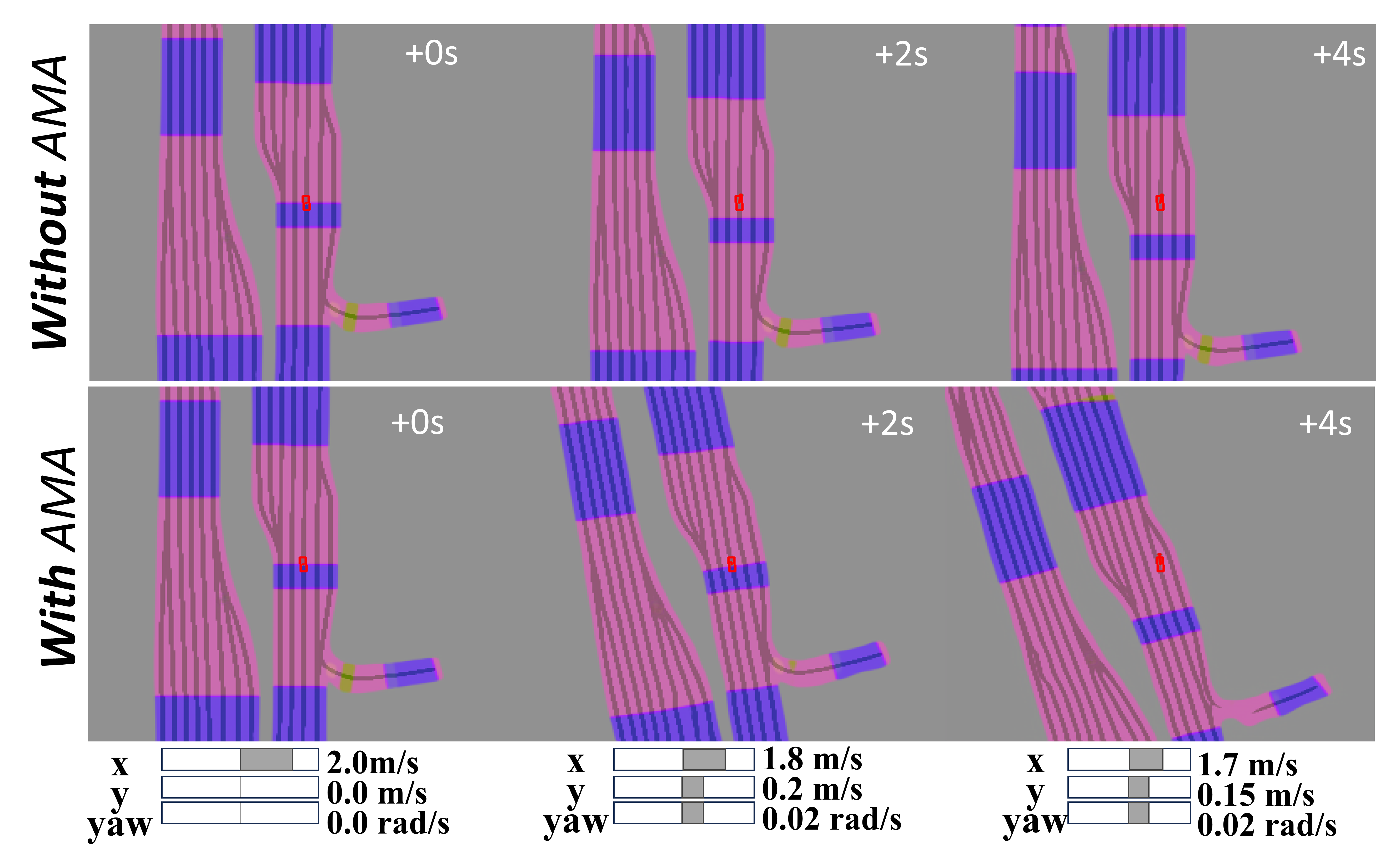}
  \vspace*{-2em}
  \caption{{\bf Map predictions with and without the AMA module}. The bar below illustrates the ego vehicle's velocity over time.
  }
  \label{fig:ab_on_map}
  \vspace*{-2em}

\end{figure}

{\bf Improved coherence by AMA module.}
Given the role of the AMA module in maintaining map consistency relative to the ego vehicle's movement, we conducted experiments to assess its impact. We trained an alternative model without this module.
As illustrated in Figure~\ref{fig:ab_on_map}, we apply a forward-right velocity to the ego-vehicle while it was traveling along a straight road. With the AMA module enabled, the map adjusts accurately to reflect the ego-vehicle’s actions. In contrast, without the AMA module, the map reveals a lack of responsiveness. This experiment underscores the importance of the AMA module in enhancing coherence between ego actions and map modalities.

\vspace{-0.1cm}
\section{Conclusion and Discussion}
\textbf{Conclusion.} We have presented \Methodname, a generative framework that enhances multimodal driving scene generation by integrating ego-action, road users, traffic maps, and images. Addressing existing limitations, \Methodname~formulates scene generation as a sequence prediction task, utilizing Temporal AutoRegressive (TAR) and Ordered AutoRegressive (OAR) modules to achieve temporal coherence and cross-modal alignment with reduced computational costs. Additionally, we introduce an Action-aware Map Alignment (AMA) module to ensure consistency between ego-action and map data by dynamically aligning map features with the ego-vehicle's movement. Experimental results have demonstrated the effectiveness of \Methodname~in generating diverse and realistic scenarios. In addition, it has been experimentally validated that \Methodname~can generate user-specified scenarios by controlling both the ego vehicle and other vehicles. This further highlights its potential as an interactive closed-loop simulator for AD systems.

\textbf{Discussion.} Building on this, to further improve the quality of the generated images, one promising approach is to incorporate generated multimodal scenes as conditions for a diffusion model~\cite{Other_diffusion_Dit, Other_diffusion_stable_diffusion}. Results demonstrating this approach are provided in the appendix.




{
    \small
    \bibliographystyle{ieeenat_fullname}
    \bibliography{main}
}

\maketitlesupplementary

\noindent In this supplementary material, we provide the following sections:
\begin{itemize}
\item \textbf{Additional visualizations} generated by the AMA module to offer deeper insights into its functionality and demonstrate how it ensures consistency between ego-action and map modalities.
\item \textbf{Detailed explanations} of the tokenization process, token embedding mechanisms, the configuration of hyperparameters for both the model architecture and the training setup, and decoder design.
\item \textbf{Comprehensive quantitative evaluation} of the generated modalities.
\item \textbf{Additional visualizations} of the comparative image quality with and without diffusion decoder 
\end{itemize}
 \vspace{-0.3cm}





\section*{A. Visualization of the transformed map in AMA module}
To better illustrate the purpose and effect of the transform operation within the AMA module, we employ visualizations to demonstrate how map features are dynamically updated in response to the ego-vehicle's motion. 

\begin{figure}[h]
  \centering
  \includegraphics[width=1\linewidth]{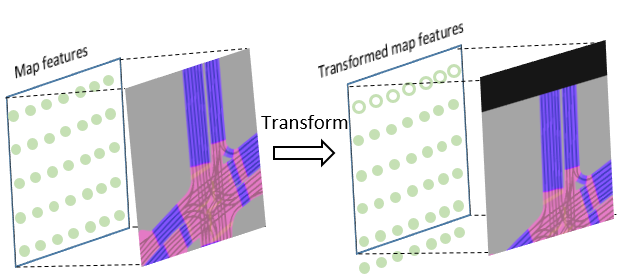}
  \caption{{\bf Visualization of the map and map features, before and after transformation.} Solid points represent existing map features, while hollow points indicate filled-in features for previously unseen regions. The ego-vehicle’s forward motion causes the map features to shift backward in the transformed map of the next frame, illustrating the alignment between the map feature and the ego-vehicle's action }
  \label{fig:map_transformed}
\end{figure}

As shown in Fig.~\ref{fig:map_transformed}, the ego-vehicle performs a forward motion at the current timestep. Consequently, in the transformed map for the next frame, the features from the current map shift backward relative to the ego-vehicle's new position. Regions previously outside the forward view are filled in to account for the unseen areas. This visualization highlights how the transform operation aligns the map features with the vehicle's actions, ensuring spatial and temporal consistency.
 \vspace{-0.3cm}

 \vspace{-0.3cm}

\section*{B. Implementation Details} 
\label{sec:Detailed framework}  

\subsection*{1. Tokenization Method}  
The tokenization standardizes and discretizes various elements of the driving scene, including ego-actions, raster maps, agents, and images. This ensures compatibility across modalities and facilitates efficient sequence generation. 

\subsubsection*{1.1 Normalization and Discretization}  
Continuous variables are normalized to the range \([0, 1]\) using:  
\begin{equation}  
v_{\text{norm}} = \frac{v - v_{\text{min}}}{v_{\text{max}} - v_{\text{min}}},  
\end{equation}  
where \(v\) is the original value, \(v_{\text{min}}\) and \(v_{\text{max}}\) are predefined bounds, and \(v_{\text{norm}}\) is the normalized result.  

After normalization, the token ID \(v_{\text{ID}}\) is assigned by discretizing the normalized value \(v_{\text{norm}}\) into one of 1024 equal intervals within the range \([0, 1]\). Each interval corresponds to a unique token ID. Specifically, the token ID is determined by identifying the interval that contains \(v_{\text{norm}}\):  
\begin{equation}  
v_{\text{ID}} = i \quad \text{such that} \quad \frac{i}{1024} \leq v_{\text{norm}} < \frac{i+1}{1024},  
\end{equation}  
where \(i \in \{0, 1, \dots, 1023\}\).

\subsubsection*{1.2 Latent Encoding for Image-like Data}  
For raster maps and images, two separate pre-trained VQ-GAN models~\cite{other_VQGAN} are used to encode the data into latent vectors, respectively. Each latent vector \(\mathbf{z}\) is quantized to the nearest codebook entry from a set of \(N+1\) learned embeddings \(\{\mathbf{e}_0, \mathbf{e}_1, \dots, \mathbf{e}_N\}\), assigning the token ID \(t\):  
\begin{equation}  
t = i \quad \text{such that} \quad \Vert \mathbf{z} - \mathbf{e}_i \Vert^2 \leq \Vert \mathbf{z} - \mathbf{e}_j \Vert^2, \quad \forall j \neq i,  
\end{equation}  
where \(i, j \in \{0, 1, \dots, N\}\).

\subsection*{2. Tokenization Process}  
The tokenization process for different modalities leverages attribute ranges derived from statistical analysis of the nuPlan dataset~\cite{agent_simu_nuplan}. Table~\ref{tab:variable_ranges} provides a summary of these ranges, which serve as the basis for consistent normalization and discretization across modalities.

\begin{table}[htbp]  
\centering  
\caption{Predefined minimum and maximum values for attribute}  
\label{tab:variable_ranges}  
\begin{tabular}{c|c|c}  
\hline  
\textbf{Attribute} & $v_{\text{min}}$ & $v_{\text{max}}$ \\  
\hline  
Position (\(x\)) & \(-64 \, \text{m}\) & \(64 \, \text{m}\) \\  
Position (\(y\)) & \(-64 \, \text{m}\) & \(64 \, \text{m}\) \\  
Position (\(z\)) & \(-5 \, \text{m}\) & \(5 \, \text{m}\) \\  
\hline  
Agent Length & \(0 \, \text{m}\) & \(15 \, \text{m}\) \\  
Agent Width & \(0 \, \text{m}\) & \(4 \, \text{m}\) \\  
Agent Height & \(0 \, \text{m}\) & \(5 \, \text{m}\) \\  
\hline  
Heading & \(-\pi \, \text{rad}\) & \(\pi \, \text{rad}\) \\  
\hline  
Speed (\(v_x\)) & \(-20 \, \text{m/s}\) & \(20 \, \text{m/s}\) \\  
Speed (\(v_y\)) & \(-20\, \text{m/s}\) & \(20 \, \text{m/s}\) \\  
Speed (\(v_z\)) & \(-0.3 \, \text{m/s}\) & \(0.3 \, \text{m/s}\) \\  
\hline  
Ego-action Displacement (\(dx\)) & \(0 \, \text{m}\) & \(10 \, \text{m}\) \\  
Ego-action Displacement (\(dy\)) & \(-0.5 \, \text{m}\) & \(0.5 \, \text{m}\) \\  
Ego-action Angular Change & \(-0.25 \, \text{rad}\) & \(0.25 \, \text{rad}\) \\  
\hline  
\end{tabular}  
\end{table}

\subsubsection*{2.1 Ego-actions}  
Ego-vehicle actions include displacements (\(x, y\)) and angular changes relative to the previous timestep. These are tokenized using the normalization and discretization process.

\subsubsection*{2.2 Raster Maps}  
A \(128 \times 128\)-meter map centered on the ego-vehicle is rasterized into a grid. Each cell represents a fixed spatial resolution and encodes one of six road types: lane, stop line, crosswalk, intersection, middle lane line, and lane connector. This produces a tensor \(\mathbf{m} \in \mathbb{R}^{256 \times 256 \times 6}\).  

A pre-trained VQ-GAN model processes the tensor, quantizing each latent vector to produce token IDs that represent the map structure.  

\subsubsection*{2.3 Agents}
 \vspace{-0.2cm}
Agents—including vehicles, pedestrians, and cyclists—are represented as 11-dimensional vectors. Attributes include position (\(x, y, z\)), speed ($v_x, v_y, v_z$), and heading, all defined relative to the ego-vehicle coordinate system, as well as dimensions (length, width, height) and category.

\begin{itemize}
    \item \textbf{Position, Speed, and Heading}: Tokenized via normalization and discretization.  
    \item \textbf{Category}: Assigned discrete IDs—1024 for vehicles, 1025 for pedestrians, and 1026 for cyclists.  
\end{itemize}  

If fewer than 64 agents are present in a scenario, \textit{pad tokens} are used. Each \textit{pad token} (ID: 1027) has default values for all attributes, ensuring they do not interfere with downstream processing.  

\begin{table}[h!]
\centering
\caption{Hyperparameter configuration of the model architecture}
\begin{tabular}{c|c}
\hline
\textbf{ Hyperparameter}            & \textbf{Value}      \\ \hline
Feature Dimension                  & 768                 \\
CA$_{\text{hist}}$ Layers in Ego-action Prediction Module & 12 \\ 
CA$_{\text{env}}$ Layers in Ego-action Prediction Module  & 12 \\ 
Temporal Causal Self-attention Layers in TAR & 24 \\
Bidirectional Self-attention Layers in TAR   & 24 \\
Causal Self-attention Layers in OAR          & 24 \\ \hline
\end{tabular}
\label{tab:hyperparameters_model}
\end{table}

\begin{table}[h!]
\centering
\caption{ Training Configuration}
\setlength{\tabcolsep}{7mm}
\begin{tabular}{c|c}
\hline
\textbf{Hyperparameter}                 & \textbf{Value}      \\ \hline
Learning Rate                      & $1 \times 10^{-4}$  \\ 
Batch Size                         & 192                 \\ 
Optimizer                          & AdamW               \\ 
Number of Training Epochs          & 300                 \\ 
Block Size                         & 20                  \\ 
Dropout Rate                       & 0.15                 \\ \hline
Temperature                        & 1.0                 \\ 
Top-$k$                            & 16                   \\ \hline
Codebook Size of Ego-action     & 1024              \\ 
Codebook Size of Raster Map     & 8192                  \\ 
Codebook Size of Agent          & 1028            \\ 
Codebook Size of Image         & 8192                  \\ \hline
\end{tabular}%

\label{tab:training_config}
\end{table}

\subsubsection*{2.4 Images}  
Input images of size \(512 \times 256\) are encoded using a pre-trained VQ-GAN. Each latent vector is quantized, and token IDs form a grid that retains the spatial and semantic structure of the original image.


\subsection*{3. Token Embedding}
To enable the model to process multimodal data effectively, tokens from different modalities are concatenated into a single sequence, which serves as the input to the embedding layer. A start token and an end token delineate the boundaries of each modality, ensuring structural clarity and facilitating positional encoding. 

Each modality is assigned a separate learnable codebook to obtain embeddings for its tokens. For the \textit{Image} and \textit{Raster map} modalities, the codebooks are initialized with pre-trained codebook weights from their respective VQ-GAN models at the start of training. Subsequently, a positional encoding is added to each token embedding, as described in the main text. This positional encoding ensures that the sequential order and spatial relationships within the tokenized data are preserved, enabling the model to capture both modality-specific features and their contextual dependencies.

\subsection*{4. Model Hyperparameters and Training Settings}
\subsubsection*{4.1 Model Structure} 
As shown in Table~\ref{tab:hyperparameters_model}, we list out the detailed hyperparameters of our model structure.

\subsubsection*{4.2 Training Setting} 
As shown in Table~\ref{tab:training_config}, we provide a detailed list of the hyperparameters used in the training setup. The block size denotes the number of timesteps or data points considered for sequence modeling. The temperature and top-\(k\) parameters control the stochasticity and diversity of the output during sampling, where temperature adjusts the probability distribution of predictions, and top-\(k\) limits the sampling to the \(k\)-most probable candidates.

\subsection*{5. Decoder Design}
\begin{figure}[h]
\centering
\includegraphics[width=1\linewidth]{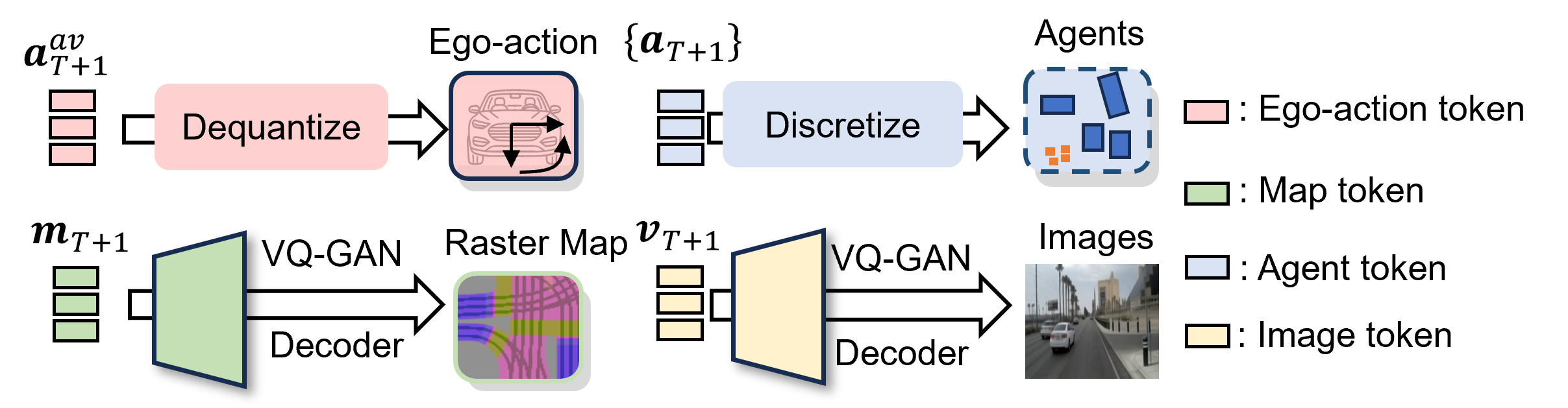}
\caption{{\bf Visualization of the Decoder.} The decoder maps ego-action and agent attribute tokens to continuous values using predefined ranges, while pre-trained VQGAN decoders reconstruct map and image tokens.}
\label{fig:decoder}
\end{figure}

The decoder transforms tokens back into their original representations. For ego-action and agent attributes, it maps tokens to continuous values based on predefined ranges (see Table.~\ref{tab:variable_ranges} in supplementary materials). For map and image tokens, pre-trained VQGAN decoders reconstruct the original data. The architecture is shown in Fig.~\ref{fig:decoder}.

\section*{C. Quantitative Evaluation of Generated Modalities} 
In this section, we present a comprehensive quantitative evaluation of the generated modalities, focusing on image quality, ego-action and agent trajectory prediction, and map realism. For image quality assessment, we employ the Fréchet Inception Distance (FID) metric, as detailed in Tab.~\ref{tab:image_quality}. To evaluate the accuracy of ego-action and agent trajectory predictions, we utilize the $\ell_2$ distance, with results summarized in Tab.~\ref{tab:L2_ego_agent}. As a baseline, we consider a scenario where both agents and the ego-vehicle maintain the velocity from the last frame, providing a reference point for comparison.

Since FID is not well-suited for evaluating multi-channel maps, we instead assess the realism of generated maps using a PatchGAN discriminator derived from our pre-trained VQGAN model. The realism score, averaged across all samples (see Tab.~\ref{tab:Map_Realism}), quantifies how convincingly the generated maps resemble real-world counterparts. A higher score indicates greater realism, while a score close to 0 suggests that the discriminator struggles to classify the map as real. 
Furthermore, we measure the distributional distance between generated map samples and ground truth (GT) map data using Maximum Mean Discrepancy (MMD), providing insights into the alignment of the generated distributions with the real data.

\begin{table}[h]
    \centering
    \small 
    \setlength{\tabcolsep}{4pt} 
    \renewcommand{\arraystretch}{1} 
    \vspace*{-0.8em}    
    \caption{FID scores of generated images for different frame lengths. The '-D' indicates the use of a diffusion image decoder.}
    \vspace*{-1em}

    \begin{tabular}{|p{1.8cm}|p{1.8cm}|p{1.8cm}|p{1.8cm}|} 
        \hline
        \centering Method & \centering 32 Frames & \centering 64 Frames &  ~~128 Frames \\ \hline
        \centering UMGen & \centering 20.91 & \centering 22.96 & ~~~~~27.50 \\ \hline
        \centering UMGen-D & \centering 15.17 & \centering 18.41 & ~~~~~21.86 \\ \hline

    \end{tabular}
    \vspace*{-1em}

    \label{tab:image_quality}
\end{table}

\begin{table}[h]
    \centering
    \small
    \vspace*{-1em}
    \setlength{\tabcolsep}{9pt} 
    \renewcommand{\arraystretch}{1} 
    \caption{Evaluation of ego-action and agent trajectory predictions.}
    \vspace*{-1em}
    \begin{tabular}{|c|cc|}
        \hline
        \textbf{Method} & \multicolumn{2}{c|}{\textbf{$\ell_2$ distance to GT (m)}} \\ \hline
        & \textbf{Ego-Action} & \textbf{Agent Trajectory} \\ \hline
        Last Frame Velocity & 0.060 & 1.53 \\ \hline
        UMGen & 0.027 & 0.54 \\ \hline
    \end{tabular}
    \vspace*{-1em}

    \label{tab:L2_ego_agent}
\end{table}

\begin{table}[h]
    \centering
    \small
    \setlength{\tabcolsep}{8.5pt} 

    \vspace*{-1.2em}
    \caption{Evaluation of generated map realism.}
    \vspace*{-1em}
    \begin{tabular}{|c|c|c|c|}
        \hline
        & Random & UMGen & Ground Truth \\ \hline
        Realism Score $\uparrow$ & -9.3 & 0.014 & 0.3692 \\ \hline
        MMD Score $\downarrow$ & 0.503 & 0.025 & 0 \\  
        \hline
    \end{tabular}
    \label{tab:Map_Realism}
    \vspace*{-1em}

\end{table}

\section*{D. Image Enhancement via a Diffusion Model} 

\begin{figure*}[h]
\centering
\includegraphics[width=1\linewidth]{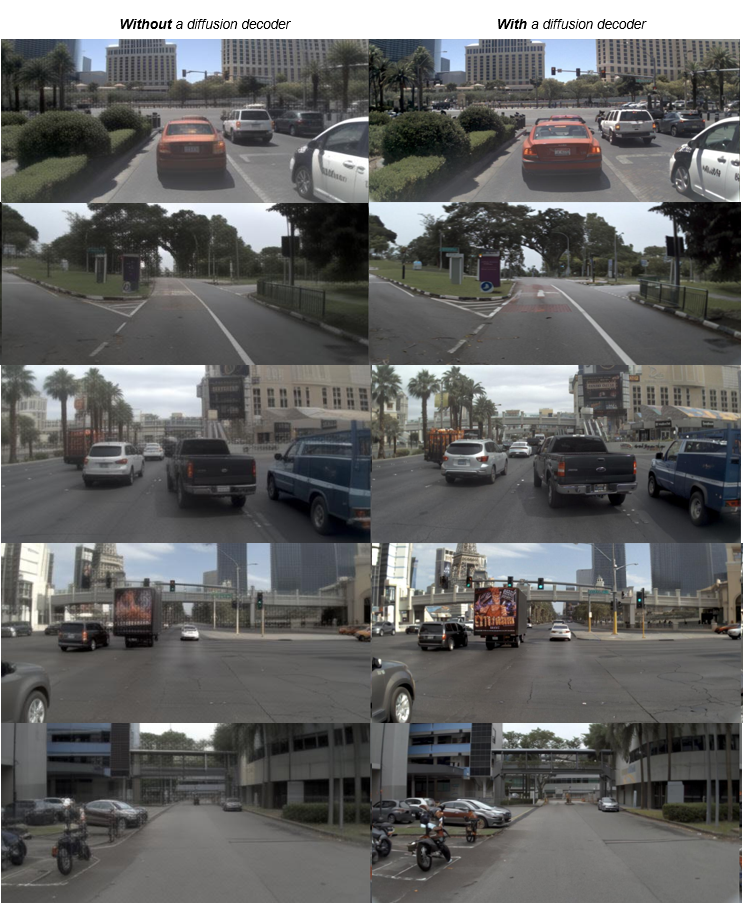}
\caption{{\bf Generated images with and without diffusion models as the image decoder. }}
\label{fig:compare_diffusion}
\end{figure*}

As discussed in the main paper, diffusion models can be a complementary part of our framework—AR ensures structured generation, while diffusion refines image fidelity. This hybrid approach has better image FID scores as in Tab.~\ref{tab:image_quality} and visual improvements in Fig.~\ref{fig:compare_diffusion}. 

\end{document}